\documentclass{article} % For LaTeX2e
\usepackage[preprint]{colm2026_conference}

\usepackage{microtype}
\usepackage{hyperref}
\usepackage{url}
\usepackage{booktabs}
\usepackage{graphicx}
\usepackage{latexsym}
\usepackage{wrapfig}
\usepackage{amsmath}
\usepackage{amssymb}
\usepackage{multirow}
\usepackage{fix-cm}
\usepackage{tcolorbox}
\usepackage{xcolor}
\usepackage[utf8]{inputenc}
\usepackage{amsfonts}
\usepackage{algorithm}
\usepackage{algpseudocode}
\usepackage{amsthm}
\usepackage{dsfont}
\usepackage{xcolor}
\usepackage{colortbl}
\usepackage{enumitem}
\usepackage{subcaption}  % 在 preamble 中

\let\cite\citep
\usepackage{lineno}

\definecolor{darkblue}{rgb}{0, 0, 0.5}
\hypersetup{colorlinks=true, citecolor=darkblue, linkcolor=darkblue, urlcolor=darkblue}

\title{Reinforcement Learning from Denoising Feedback}

\author{
    Qi He\textsuperscript{1,2}\thanks{Work done during internship at Ant Group}, \ 
    Huan Chen\textsuperscript{2}, \ 
    Ya Guo\textsuperscript{2}, \
    Huijia Zhu\textsuperscript{2}, \ 
    Yi R. (May) Fung\textsuperscript{3}, \ 
    Baojian Zhou\textsuperscript{1} \\[6pt]
    \textsuperscript{1}\,Fudan University
    \textsuperscript{2}\,Ant Group \\
    \textsuperscript{3}\,Hong Kong University of Science and Technology \\[4pt]
    \texttt{qhe22@m.fudan.edu.cn,\ bjzhou@fudan.edu.cn}\\
    \texttt{\{chenhuan.ch, guoya.gy, huijia.zhj\}@antgroup.com} \\
    \texttt{yrfung@cse.ust.hk}
}

\begin{document}

\ifcolmsubmission
\linenumbers
\fi

\maketitle

\begin{abstract}
Policy loss estimation remains a fundamental and long-standing challenge in reinforcement learning (RL) for diffusion language models (DLMs). We introduce \textbf{Reinforcement Learning from Denoising Feedback (RLDF)}, a novel training paradigm that leverages feedback obtained from rollout and training processes to facilitate accurate and efficient policy loss estimation. To balance the trade-off between computational efficiency and estimation effectiveness, RLDF optimizes the model toward the clipped clean state from intermediate noisy states, combined with weighted timestep sampling over denoising timesteps. Extensive experiments demonstrate that RLDF achieves consistent and substantial improvements in both performance and generalizability across two representative DLM architectures, LLaDA and Dream, on multiple reasoning benchmarks. Our work lays a principled foundation for scalable reinforcement learning in diffusion language models. We build \textbf{Drift}, a training framework for DLMs, available at \url{https://github.com/ant-research/Drift}.
\end{abstract}

\section{Introduction}

Recent breakthroughs in diffusion language models (DLMs)~\cite{austin2021structured,sahoo2024simple}, such as Dream~\cite{ye2025dream} and LLaDA~\cite{nie2025large}, have positioned them as a promising pathway toward highly capable foundation models~\cite{googledeepminddiffusion2025,song2025seed,khanna2025mercury}. The appeal of DLMs~\cite{gulrajani2023likelihood,yu2025discrete} stems from several intrinsic advantages over autoregressive (AR) architectures: bidirectional attention~\cite{gong2025diffucoder}, arbitrary decoding order, and the potential for multi-token parallel generation~\cite{wu2025fast,wu2025fastv2}. Nevertheless, DLMs currently underperform their AR counterparts across many benchmarks, a gap largely attributable to immature training methodologies and pipelines. Among the capabilities most in need of improvement, reasoning~\cite{openai2024o1,anthropic2025claude37,guo2025deepseek,team2025kimi} stands out as particularly critical. Reasoning of language models is a frontier that has been extensively explored and substantially advanced within the autoregressive paradigm.

Reinforcement learning (RL)~\cite{kaelbling1996reinforcement} has emerged as an effective and highly generalizable approach for cultivating reasoning capabilities in large language models (LLMs). Evolving from PPO~\cite{schulman2017proximal} through DPO~\cite{rafailov2023direct}, GRPO~\cite{shao2024deepseekmath}, and subsequent variants~\cite{yu2025dapo,yue2025vapo}, RL has consistently improved LLM reasoning with verifiable reward~\cite{wang2024math,uesato2022solving} across diverse scenarios, including tool use~\cite{jin2025search}, mathematics~\cite{snell2024scaling}, and code generation~\cite{le2022coderl,shojaee2023execution}. However, these RL methods cannot be directly transplanted to diffusion language models due to two fundamental discrepancies: the attention mechanism (causal vs. bidirectional) and the generative process (next-token prediction vs. full-sequence denoising).

A central obstacle in adapting RL to DLMs lies in \textbf{policy loss estimation}. In autoregressive models, the policy loss is obtained through a single forward pass under causal attention. In contrast, the arbitrary decoding order and bidirectional attention of DLMs render exact log-likelihood computation intractable. Prior approaches~\cite{zhao2025d1,ou2025principled} approximate the policy loss via either all-masking or random-masking schemes, both of which introduce a substantial mismatch between training and inference. Alternatively, computing the policy loss at every denoising step~\cite{ni2026flexibility} preserves train–inference alignment but incurs prohibitive computational and time costs due to the iterative nature of the denoising process. \textbf{This exposes a core challenge: how to balance the fidelity of policy loss estimation against its computational cost.}

% To address this challenge, we propose \textbf{Reinforcement Learning from Denoising Feedback (RLDF)}, a training paradigm that leverages signals naturally produced during the DLM inference process. Specifically, we exploit the decoding step at which each token is committed and its associated decoding probability as criteria for training-step selection. Building on this signal, we employ a weighted sampling strategy~\cite{rubinstein2016simulation} to preferentially sample denoising steps at which the model exhibits high uncertainty~\cite{choi2022perception}—steps that exert the greatest influence on the final output. The policy loss is then computed only on these selected steps and used as the training signal for policy updates. Our method achieves a favorable trade-off between precision and efficiency, while remaining compatible with diverse masking strategies and readily extensible to novel training schemes.
To address this challenge, we propose \textbf{Reinforcement Learning from Denoising Feedback (RLDF)}, a training paradigm that leverages two complementary feedback sources. \textbf{Rollout feedback} captures unmasking probabilities and their corresponding timesteps along the denoising trajectory. \textbf{Training feedback}, derived from predicted probabilities during denoising training, provides signals from which we retain only the most informative positions for stable and effective policy updates. Building on these signals, we employ a weighted sampling strategy~\cite{rubinstein2016simulation} that preferentially selects denoising steps with high predictive uncertainty, exerting the greatest influence on the final generation~\cite{choi2022perception}. The policy loss is computed exclusively over these selected steps, yielding a favorable precision–efficiency trade-off while remaining compatible with diverse masking strategies and readily extensible to emerging training schemes.

Empirical results demonstrate that our method exhibits substantial improvements on math and code tasks across both LLaDA and Dream models, as shown in Table~\ref{tab:main_result}. Furthermore, our approach achieves faster convergence rates with more precise policy gradient estimation. We also demonstrate the advantage of leveraging the clean state rather than the next state for loss estimation, which avoids loss spikes and improves the utilization of training signals. In summary, our contributions can be summarized as follows:

\begin{itemize}[leftmargin=*, itemsep=2pt, parsep=0pt, topsep=0pt]
    \item We open-source \textbf{Drift}, an extensible and modular RL framework purpose-built for DLMs, to facilitate community research and novel algorithms.
    \item We propose an improved policy loss estimation method that achieves a better trade-off between efficiency and precision.
    \item We unlock the reasoning potential of DLMs (LLaDA and Dream), providing empirical insights for DLM community.
\end{itemize}
% \begin{itemize}[topsep=0pt, itemsep=2pt, parsep=0pt]
%     \item We open-source an extensible and modularized reinforcement learning framework to advance the development of DLM training.
%     \item We propose an improved policy loss estimation method that achieves a better trade-off between efficiency and precision.
%     \item We unlock the reasoning potential of DLMs (LLaDA and Dream), providing empirical insights for DLM community.
% \end{itemize}

\section{Related Work}

\subsection{Diffusion Language Model}
Diffusion language models (DLMs) have evolved significantly, spanning approaches from continuous~\cite{li2022diffusion,dieleman2022continuous} to discrete domains~\cite{hoogeboom2021argmax}. Recently, masked diffusion language models (MDLMs) define a text generation paradigm in which the output is produced through an iterative denoising process over discrete token sequences~\cite{austin2021structured}. In contrast to autoregressive models, which generate tokens sequentially from left to right, DLMs employ bidirectional attention mechanisms and perform predictions across all token positions at each forward pass. Recent advances in training methodology have yielded a number of capable DLM foundation models, including LLaDA~\cite{nie2025large}, which is trained from scratch via full pre-training, and Dream~\cite{ye2025dream}, which is obtained by adapting the pretrained Qwen2.5-7B~\cite{yang2025qwen3} to the diffusion framework.

\subsection{Reasoning in Diffusion Language Models}
Reasoning has long been a central topic in language model research. DLMs have recently demonstrated promising potential across a range of reasoning tasks. The models show superiority in planning tasks~\cite{ye2024beyond,huang2025reinforcing} and demonstrate better data learning efficiency~\cite{ni2025diffusion,prabhudesai2025diffusion,gao2025makes}. Complementary lines of work have explored augmenting reasoning capabilities through established techniques such as chain-of-thought prompting~\cite{ye2024diffusion}, leveraging both supervised fine-tuning and reinforcement learning~\cite{ho2020denoising,austin2021structured} to elicit more structured and reliable inference. Further efforts have focused on accelerating the reasoning process~\cite{xie2025dream}, while others have investigated the design of remasking strategies to improve generation quality~\cite{cheng2026reasoning}.

\subsection{Reinforcement Learning}

Reinforcement learning (RL)~\cite{kaelbling1996reinforcement,schulman2017proximal}, especially GRPO~\cite{shao2024deepseekmath}, has proven effective in enhancing the reasoning capabilities of large language models (LLMs)~\cite{he2025veri,qian2025toolrl}. However, owing to fundamental differences in attention mechanisms and forward computation processes, diffusion language models require distinct strategies for computing RL losses. Existing loss computation methods have undergone successive refinements, evolving from Diffu-GRPO~\cite{zhao2025d1,tang2025wd1} through Coupled-GRPO~\cite{gong2025diffucoder} to ELBO~\cite{ou2025principled}, etc. Each of these methods attempts to better align the training objective with the denoising nature of diffusion models. Nevertheless, all of these methods suffer from training-inference mismatch to varying degrees, which undermines the reliability of policy gradient estimation and ultimately limits reasoning performance. We are motivated by this limitation and propose a novel loss computation method that considers both accuracy and efficiency, serving as a stronger foundation for RL on DLMs.

\section{Preliminaries}
\subsection{Forward and Denoising Process}

The forward process independently masks each token $x_{0,i}$ with probability $t$ or leaves it unchanged with probability $1-t$, yielding the partially masked sequence $x_t$:
\begin{equation}
    q(x_t \mid x_0) = \prod_{i=1}^{L} \mathrm{Cat}\!\left(x_{t,i};\; (1-t)\,\mathbf{e}_{x_{0,i}} + t\,\mathbf{e}_{\texttt{[MASK]}}\right).
\end{equation}
During denoising, DLMs support both static and dynamic strategies. In static decoding, the model unmasks the top-$k$ most confident token positions at each timestep, where $k$ is a fixed hyperparameter. In dynamic decoding, tokens are unmasked based on predicted logits, with positions whose confidence exceeds a predefined threshold being unmasked.

To precisely describe the denoising process, we formalize the denoising process with the following definition.
Given a prompt $q$, let $\{o_T, o_{T-1}, \ldots, o_0\}$ describe the denoising process, where $T$ is the number of denoising steps at inference time. Each $o_t$ represents the model's response state at denoising timestep $t$. Specifically, $o_T$ corresponds to the fully masked initial state, 
$o_0$ denotes the final clean response, and intermediate states $o_t$ ($0 < t < T$) 
contain a mixture of decoded tokens and remaining mask tokens. 

\subsection{Training Techniques}

Following the masked language modeling paradigm of BERT, MDLMs are trained via random masking. Specifically, a parametric model $p_\theta(\cdot \mid x_t)$ takes the partially masked sequence $x_t$ as input and simultaneously predicts all masked tokens (denoted as $\mathtt{M}$). It is optimized via a cross-entropy loss computed exclusively over the masked positions:
\begin{equation}
    \mathcal{L}(\theta) \triangleq -\mathbb{E}_{t,\, x_0,\, x_t}\left[\frac{1}{t}\sum_{i=1}^{L}\mathds{1}[x_{t,i} = \mathbf{M}]\log p_\theta(x_{0,i} \mid x_t)\right].
\end{equation}
However, a fundamental discrepancy exists between the training and denoising processes. During training, masking is applied randomly and independently across token positions, whereas during denoising, the set of masked positions evolves deterministically across steps based on model confidence. This mismatch results in substantially different masking patterns and contextual inputs between the two phases, leading to inconsistent attention distributions and key-value representations. Such train-inference discrepancy poses a significant challenge for reasoning-oriented fine-tuning, as the model is optimized under a distribution of masked contexts that poorly reflects those encountered at inference time.

\subsection{Challenges in Log Probability Estimation}
Unlike autoregressive models, whose loss can be computed within a single forward pass, diffusion language models do not enjoy this benefit due to their bidirectional attention mechanism. Specifically, the key-value (KV) representations change across denoising steps, making the forward computation distinct at each step. Early methods such as d1~\cite{zhao2025d1} and wd1~\cite{tang2025wd1} adopt a full-sequence loss strategy, which computes the average loss over all token positions conditioned on the prompt only in a single forward pass:
\begin{equation}
    \mathcal{L}_{\text{full-seq}}(o\mid q) = \log p_\theta(o \mid q, o_T).
\end{equation}
Subsequently, the Evidence Lower Bound (ELBO) was introduced, which approximates the log-likelihood via random masking over denoising timesteps:
\begin{equation}
    \mathcal{L}_{\text{ELBO}}(o \mid q) = 
    \mathbb{E}_{t,\, m}\left[ 
    \log p_\theta\big(o \mid q,\, o^{m}\big)\right],
\end{equation}
where $o^m$ denotes the response sequence randomly masked by $\mathcal{M}$.

However, all of these methods compute the loss under a context that is inconsistent with the actual rollout conditions at inference time, giving rise to a fundamental training-inference mismatch that undermines the reliability of policy gradient estimation in RL training.

\section{Reinforcement Learning from Denoising Feedback}

\begin{figure*}[!t]
    \centering
    \includegraphics[width=\textwidth]{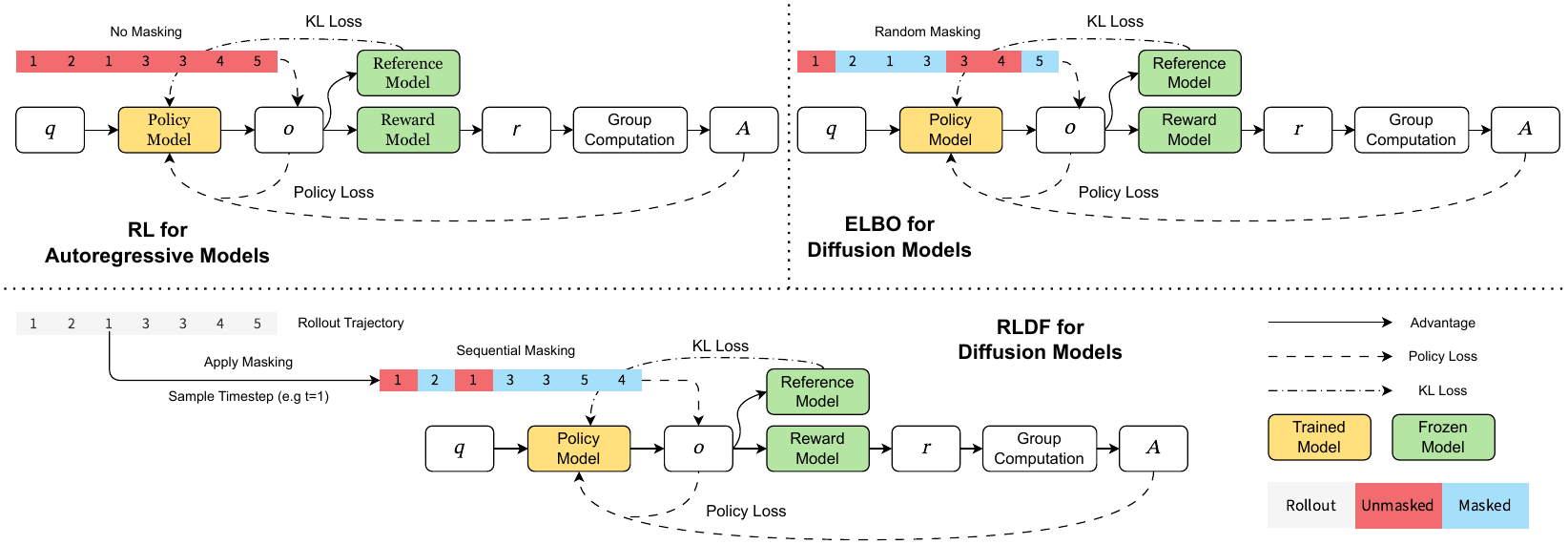}
    \caption{Comparison of training paradigms. 
        \textbf{Left:} RL for autoregressive models without masking. 
        \textbf{Right:} ELBO-based training for diffusion models with random masking. 
        \textbf{Bottom:} Our proposed RLDF, which samples a timestep t and applies sequential masks. Policy loss is estimated based on $\pi_{\theta}(x_0 \mid x_t)$. This formulation provides a more precise policy loss estimation while using identical trajectory context and objective.}
    \label{fig:rldf-framework}
\end{figure*}

\subsection{Denoising Feedback of Diffusion Models}
The denoising process of diffusion models contains rich information that can substantially benefit reinforcement learning. We extract feedback from two complementary stages: the \textbf{rollout process} and the \textbf{training process}. During rollout, we collect reasoning trajectories together with their timestep probabilities, which capture the reasoning context and guide the timestep selection. During training, we predict the clean state $x_0$ from the noisy state $x_t$ and derive the learning signal from the processed $\hat{x}_0$ through token-level clipping. Both sources of feedback are effective in stabilizing training, yet have been largely overlooked in prior paradigms.

\subsection{Loss Estimation at Timestep t}

\begin{figure}[!t]
    \centering
    \begin{subfigure}[t]{0.49\columnwidth}
        \centering
        \includegraphics[width=\linewidth]{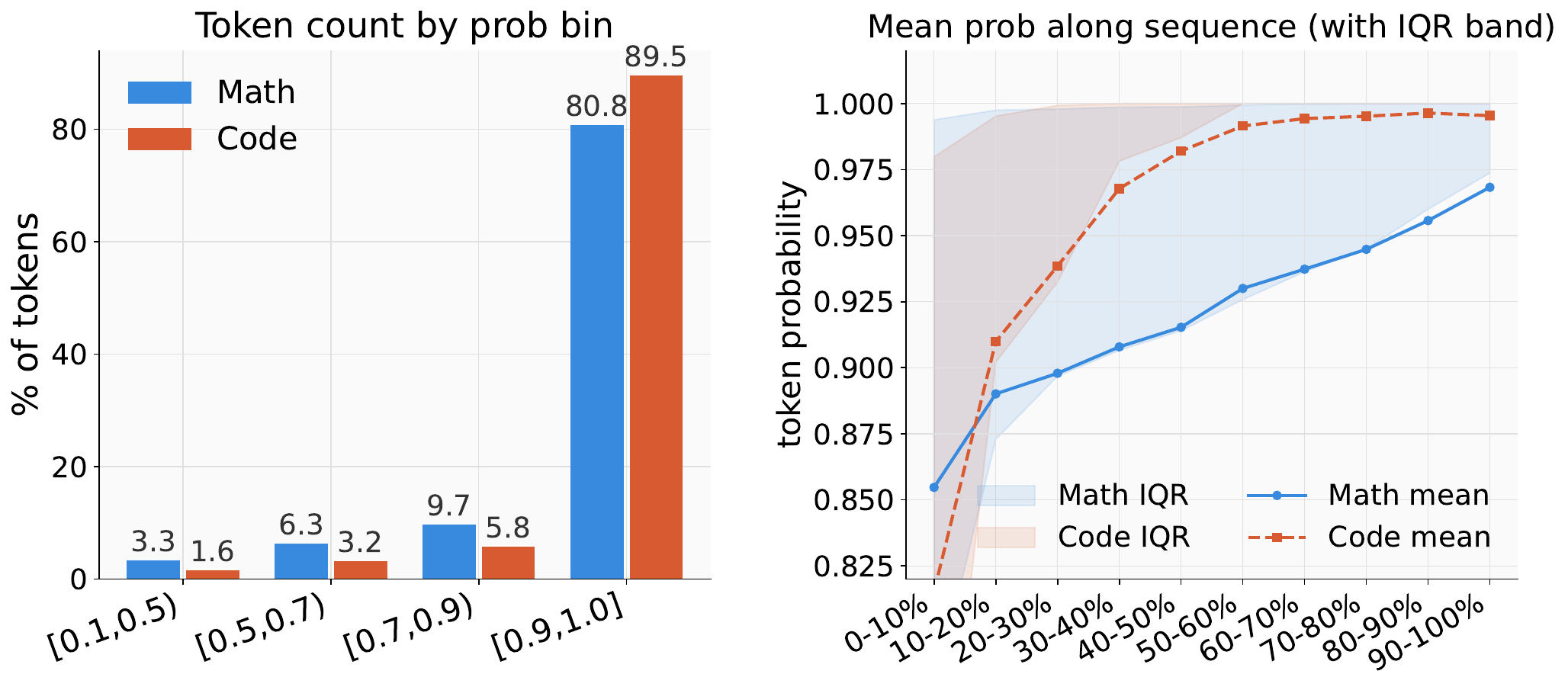}
        \caption{Token probability distributions on two tasks. 
            \textit{(left)} Over 80\% of tokens are unmasked with high conf.\ ($p \geq 0.9$) in both tasks. 
            \textit{(right)} Mean token probability along denoising timesteps (IQR shaded); later-unmasked tokens show higher conf.\ and lower var.}
        \label{fig:token-prob}
    \end{subfigure}
    \hfill
    \begin{subfigure}[t]{0.49\columnwidth}
        \centering
        \includegraphics[width=\linewidth]{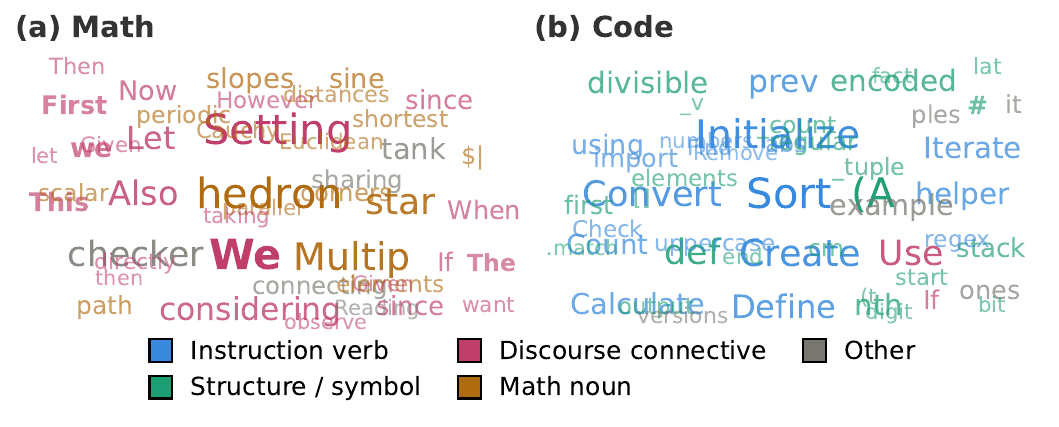}
        \caption{Word clouds of high-entropy tokens during decoding on Math (left) and Code (right) tasks. Font size encodes mean entropy $H$; font weight encodes frequency (bold: $\geq$50, medium: 10--49, regular: $<$10). Colors denote semantic category.}
        \label{fig:wordcloud}
    \end{subfigure}
    \caption{\textbf{Token-level analysis during denoising process.} (a) Per-token entropy distributions across Math and Code tasks; (b) word clouds of high-entropy tokens.}
    \label{fig:token-analysis}
\end{figure}

To faithfully replicate the context in which each position is unmasked, we adopt a stepwise loss estimation scheme that enables precise gradient estimation without approximation. Formally, given a prompt $q$ and the sequence of tokens decoded across denoising timesteps $o_T, \ldots, o_0$, at each timestep $t$ we compute the forward process $p(o_0 \mid q, o_t)$ and evaluate the loss exclusively on the positions masked at timestep $t$. This procedure exactly mirrors the model's rollout without introducing any approximation error.

However, this exact formulation incurs substantial computational overhead. Specifically, the number of forward passes scales linearly with the number of decoding steps $n$, as the loss must be computed iteratively for each step. In practice, a standard reasoning trajectory requires approximately 256 decoding steps, resulting in 256 forward passes per sample and rendering naive stepwise training prohibitively expensive in both time and memory.

\subsection{Weighted Sampling for Timestep Selection}

To mitigate the computational bottleneck, we propose selectively sampling a subset of denoising steps for loss computation rather than processing all steps exhaustively, adopting weighted sampling as the principled mechanism for timestep selection.

Prior work has shown that entropy serves as an effective guidance signal during training~\cite{wang2025beyond,cheng2026reasoning}. As shown in Table~\ref{tab:correlation}, distribution entropy and token probability at each position exhibit near-perfect correlation across both models and tasks, indicating that decoding probability is a faithful proxy for entropy. We therefore adopt per-step decoding probability as the weight for timestep selection, which not only provides a more intuitive perceptual signal but also avoids explicit entropy computation.

\begin{wraptable}{r}{0.45\columnwidth}
\vspace{-1em}
\centering
\small
\resizebox{0.45\columnwidth}{!}{%
\begin{tabular}{llcc}
\toprule
\textbf{Model} & \textbf{Task} & \textbf{Pearson} $r$ & \textbf{Spearman} $\rho$ \\
\midrule
\multirow{2}{*}{LLaDA} & Math & $-0.942$ & $-0.994$ \\
                        & Code & $-0.950$ & $-0.933$ \\
\addlinespace[2pt]
\multirow{2}{*}{Dream}  & Math & $-0.930$ & $-0.994$ \\
                        & Code & $-0.928$ & $-0.997$ \\
\bottomrule
\end{tabular}}
\caption{Pearson ($r$) and Spearman ($\rho$) correlations between each token's unmasking probability and entropy of its predicted distribution, computed on LLaDA and Dream across MATH and MBPP.}
\label{tab:correlation}
\vspace{-1em}
\end{wraptable}

As illustrated in Figure~\ref{fig:token-prob}, the distribution of per-token decoding probabilities is heavily skewed: the vast majority of tokens are decoded with high confidence (typically above 0.9), while only a small fraction are assigned low probability. Crucially, these low-confidence tokens, though sparse, exert disproportionate influence on the diversity of reasoning trajectories and exhibit strong correlation with final answer accuracy. As shown in Figure~\ref{fig:wordcloud}, such tokens are often connective words in math and instruction verbs in code tasks. We therefore preferentially sample these informative steps with lower average token confidence for loss computation.

For diffusion language models, this sampling strategy offers a two-fold benefit: it substantially reduces the number of loss computation iterations, while simultaneously concentrating gradient signal on the most informative denoising steps, yielding a lower-variance and more rollout-faithful training objective.

\subsection{Loss Estimation of RLDF}

The RLDF loss estimation pipeline, illustrated in Figure~\ref{fig:rldf-framework}, follows a structured sequence:
rollout generation, sequential masking schedule construction, iterative loss estimation on each selected timestep, and a final aggregated backpropagation.
 
\paragraph{Sampling timesteps with probability-based weighting.}
Rather than selecting timesteps uniformly, we weight each timestep by the average token-level uncertainty of the model. Concretely, for each timestep $t$ we compute
\begin{equation}
\label{eq:step_weight}
u_t = -\frac{1}{|\mathcal{I}_t|}\sum_{i\in \mathcal{I}_t}\log p_i,
\end{equation}
where $\mathcal{I}_t$ denotes the set of token positions unmasked at timestep $t$, formally satisfying $\mathcal{I}_t \triangleq \{i : o_{t,i} \neq o_{t-1,i}\}$.
$p_i$ is the model's predicted probability for token $i$, and $u_t$ is the resulting uncertainty estimate for timestep $t$. Timesteps with lower model confidence (smaller $p_i$) thus receive larger weight. Here, $o_{t,i}$ denotes the token at position $i$ of the response at timestep $t$.
A temperature-scaled softmax converts these weights into a sampling distribution:
\begin{equation}
\label{eq:step_softmax}
w_t = \frac{\exp(u_t/\tau)}{\sum_{t'}\exp(u_{t'}/\tau)},
\end{equation}
and $k$ steps are drawn without replacement from $w$. As $\tau\!\to\!0$,
this degenerates to deterministic top-$k$ selection, while
$\tau\!\to\!\infty$ recovers uniform sampling, allowing a smooth
interpolation between exploitation and exploration.
 
\paragraph{Iterative loss estimation on each timestep.}
The loss at timestep $t$ of response $b$ is computed by Eq.~(\ref{eq:ppo_loss}), which reduces to Eq.~(\ref{eq:reinforce_loss}) when only a single iteration is performed. Here, $\hat{A}^{(b)}$ denotes the normalized advantage of response $b$, and $\hat{o}_0$ denotes its clipped version, which retains only tokens with predicted probability above a threshold. This clipping prevents low-confidence tokens from dominating the training signal. The per-step loss is normalized by the number of retained tokens, aggregated over the selected steps $S_t^{(b)}$, and finally averaged across all responses in the group to form the final objective.
 
When more than one forward-pass iteration is performed per rollout, the
behavior policy $\pi_{\theta_{\text{old}}}$ used to generate samples
diverges from the current policy $\pi_{\theta}$ being optimized. To correct
for this off-policy mismatch and stabilize training, we adopt a weighted sampling ratio together with PPO-style clipping:
\begin{equation}
\small
\label{eq:ppo_loss}
\ell_{\text{policy}}^{(b)}(t) = -\frac{1}{|\hat{o}_0|}\sum_{i\in \hat{o}_0}
\min\!\Big(
  r_{t, i}^{(b)}(\theta)\,\hat{A}^{(b)},\,
  \mathrm{clip}\big(r_{t, i}^{(b)}(\theta),\,1{\pm}\varepsilon\big)\hat{A}^{(b)}
\Big),\;
\text{where}\;
r_{t,i}(\theta) = \frac{\pi_{\theta}(\hat{o}_{0,i} \mid q,\, o_{t})}{\pi_{\theta_{\text{old}}}(\hat{o}_{0,i} \mid q,\, o_{t})}.
\end{equation}
When only one iteration is performed,
$\pi_{\theta}=\pi_{\theta_{\text{old}}}$, so $r_t^{(i)}(\theta)\equiv 1$
and the clipping is inactive; Eq.~(\ref{eq:ppo_loss}) then reduces to the REINFORCE form~\cite{williams1992simple}:
\begin{equation}
\label{eq:reinforce_loss}
\ell_{\text{policy}}^{(b)}(t) = -\frac{1}{|\hat{o}_0|}\sum_{i\in \hat{o_0}}
\log \pi_{\theta}(\hat{o}_{0, i}^{(b)} \mid q,\, o_{t}^{(b)})\,\hat{A}^{(b)}.
\end{equation}

\paragraph{Regularization and aggregation.}
To further constrain the policy from drifting too far from the reference
model $\pi_{\text{ref}}$ and to ensure training stability, we additionally
incorporate the $k_3$ KL-divergence estimator
~\citep{schulman2020kl} between $\pi_{\theta}$ and $\pi_{\text{ref}}$:
% {
% \begin{equation}
% \ell_{\mathrm{KL}} = \mathbb{D}_{\mathrm{KL}}[\pi_\theta \,\|\, \pi_{\mathrm{ref}}] = \mathbb{E}\left[\frac{\pi_{\mathrm{ref}}}{\pi_\theta} - \log\frac{\pi_{\mathrm{ref}}}{\pi_\theta} - 1\right]
% \end{equation}
% }
% adapted into our setting:
\begin{equation}
\ell_{\mathrm{KL}}^{(b)}(t) = \mathbb{D}_{\mathrm{KL}}[\pi_\theta(\hat{o}^{(b)}_0 \mid q,o^{(b)}_{t}) \,\|\, \pi_{\mathrm{ref}}(\hat{o}^{(b)}_0 \mid q,o^{(b)}_{t})].
\end{equation}
After computing Policy and KL loss for every selected step, we aggregate across
$S_t^{(b)}$ and then across all $G$ responses in the group to obtain the
final loss objective:
\begin{equation}
\label{eq:policy_obj}
\mathcal{L}_{\text{RLDF}} = \frac{1}{G}\sum_{b=1}^{G}
\frac{1}{|S_t^{(b)}|}\sum_{t\in S_t^{(b)}} \left[\ell_{\text{policy}}^{(b)}(t) + \beta\ell_{KL}^{(b)}(t)\right],
\end{equation}where $\beta$ controls the KL regularization strength. We additionally apply gradient-norm clipping during backpropagation to suppress occasional large updates and stabilize training. The complete training objective is given in Eq.~(\ref{eq:rldf}).
\begin{equation}
\label{eq:rldf}
\resizebox{\linewidth}{!}{$
\begin{aligned}
\mathcal{L}_{\text{RLDF}}(\theta) = \mathbb{E}_{q\sim\mathcal{D},\,o\sim\pi_{\theta_{\mathrm{old}}}(\cdot\mid q)}\Bigg[
&-\frac{1}{G}\sum_{b=1}^{G}
\mathrm{softmax}_t\bigg(
-\frac{1}{\tau\,|\mathcal{I}_t|}\sum_{i\in \mathcal{I}_t}
\log\phi_{\pi_{\theta_{\mathrm{old}}}}(\hat{o}_{t-1, i}\mid q, o_{t})
\bigg)
\frac{1}{|\hat{o}^{(b)}_0|}\sum_{i\in \hat{o}^{(b)}_0}
\bigg[\min\Big(
\frac{\phi_{\pi_\theta}(\hat{o}_{0,i}^{(b)}\mid q, o_{t}^{(b)})}
     {\phi_{\pi_{\theta_{\mathrm{old}}}}(\hat{o}_{0,i}^{(b)}\mid q, o_{t}^{(b)})}\hat{A}^{(b)},\\[2pt]
& \mathrm{clip}\big(
\frac{\phi_{\pi_\theta}(\hat{o}_{0,i}^{(b)}\mid q, o_{t}^{(b)})}
     {\phi_{\pi_{\theta_{\mathrm{old}}}}(\hat{o}_{0,i}^{(b)}\mid q, o_{t}^{(b)})},\,
1{-}\varepsilon,\,1{+}\varepsilon\big)\hat{A}^{(b)}
\Big)
+\, \beta\, D_{\mathrm{KL}}\Big[\phi_{\pi_\theta}(\cdot\mid q, o_{t}^{(b)})\,\big\|\,\phi_{\pi_{\mathrm{ref}}}(\cdot\mid q, o_{t}^{(b)})\Big]\bigg]
\Bigg]
\end{aligned}
$}
\end{equation}

\section{Experiments}
\subsection{Experiment Setup}
\paragraph{Data and Models.}
Mathematical reasoning and code generation are widely adopted benchmarks for evaluating the reasoning capabilities of large language models. Accordingly, we conduct training and evaluation on both domains. For mathematical reasoning, we use the official training split of MATH~\cite{hendrycks2021measuring} for training, and evaluate on MATH500~\cite{lightman2023let}, GSM8K~\cite{cobbe2021training}, and AMC23~\cite{aops2023amc}. For code generation, we train on the training split of MBPP~\cite{austin2021program} and evaluate on MBPP and HumanEval~\cite{chen2021evaluating}. We adopt two mainstream diffusion language models as our backbones: LLaDA-8B-Instruct\cite{nie2025large} and Dream-7B-Instruct~\cite{ye2025dream}.

\paragraph{Training.}
Our training procedure consists solely of reinforcement learning, initialized from the pretrained diffusion backbones. For policy loss estimation, we apply weighted sampling over 16 denoising timesteps. To prevent excessive deviation from the reference model, we adopt the widely used $k_3$ estimator for the KL regularization term. During rollout, we follow ~\citet{wang2025revolutionizing} and employ block-wise decoding with KV-cache to accelerate generation. Full hyperparameter and implementation details are provided in Appendix~\ref{app:impl}.

\paragraph{Evaluation.}
To highlight the generalization ability of each RL method, we adopt a strict cross-setting evaluation protocol: each model is trained on a single dataset with a fixed generation length of 256, and then evaluated across all remaining settings. Following prior work, we benchmark all models at generation lengths of 128, 256, and 512 under the dynamic and static sampling strategy. We compare our method against several well-established baselines, including 
d1~\citep{zhao2025d1}, 
ESPO~\citep{ou2025principled}, 
TraceRL~\citep{wang2025revolutionizing}, 
and Coupled-GRPO~\citep{gong2025diffucoder}. 
In addition, for the LLaDA series, we report the performance of LLaDA-1.5~\citep{zhu2025llada} as an additional reference point.

\subsection{Main Results}
\label{sec:main_results}
\begin{table*}[!htbp]
\centering
\renewcommand{\arraystretch}{1.15}
\resizebox{\textwidth}{!}{%
\begin{tabular}{l|ccc|ccc|ccc|ccc|ccc}
\toprule
\multirow{2}{*}{\textbf{Model}} & \multicolumn{3}{c|}{\textbf{MATH500}} & \multicolumn{3}{c|}{\textbf{GSM8K}} & \multicolumn{3}{c|}{\textbf{AMC23}} & \multicolumn{3}{c|}{\textbf{HumanEval}} & \multicolumn{3}{c}{\textbf{MBPP}} \\
\cmidrule(lr){2-4} \cmidrule(lr){5-7} \cmidrule(lr){8-10} \cmidrule(lr){11-13} \cmidrule(lr){14-16}
 & 128 & 256 & 512 & 128 & 256 & 512 & 128 & 256 & 512 & 128 & 256 & 512 & 128 & 256 & 512 \\
\midrule
\multicolumn{16}{l}{\textbf{LLaDA Models}} \\
\midrule
LLaDA-8B-Instruct & 32.0 & 36.4 & 40.8 & 75.7 & 83.4 & 83.0 & 7.5  & 12.5 & 12.5 & 39.0 & 37.2 & 35.9 & 37.4 & 36.6 & 38.0 \\
LLaDA-1.5         & 32.2 & 40.0 & 39.8 & 76.1 & 82.0 & 84.0 & 15.0 & 12.5 & 7.5  & 40.8 & 35.9 & 38.4 & 36.2 & 38.6 & 38.6 \\
d1  & 32.6 & 36.8 & 38.2 & 75.9 & 82.5 & 82.5 & 12.5 & \textbf{17.5} & 12.5 & 38.4 & 36.5 & 37.8 & 38.0 & 39.2 & 40.8 \\
ESPO   & 37.8 & 38.0 & 40.6 & 79.1 & 83.4 & 84.2 & 12.5 & 15.0 & \textbf{22.5} & 41.4 & 37.2 & \textbf{40.2} & \textbf{42.2} & 43.8 & 42.6 \\
TraceRL& 31.8 & 33.4 & 33.6 & 77.7 & 80.7 & 81.5 & 7.5  & 5.0  & 20.0 & 38.4 & 35.3 & 35.9 & 36.8 & 39.0 & 40.2 \\
Coupled-GRPO       & 33.2 & 41.2 & 37.2 & 77.8 & 83.5 & 83.0 & 15.0 & 10.0 & 20.0 & 37.8 & 40.9 & 39.6 & 38.6 & 42.4 & 42.0 \\
RLDF (ours)       & \textbf{39.6} & \textbf{43.8} & \textbf{45.0} & \textbf{79.2} & \textbf{84.7} & \textbf{84.9} & \textbf{15.0} & \textbf{17.5} & 20.0 & \textbf{42.8} & \textbf{40.9} & 38.5 & 40.4 & \textbf{44.6} & \textbf{44.8} \\
\rowcolor{gray!15}
$\Delta$ vs. Base & +7.6 & +7.4 & +4.2 & +3.5 & +1.3 & +1.9 & +7.5 & +5.0 & +7.5 & +3.8 & +3.7 & +2.6 & +3.0 & +8.0 & +6.8 \\
\midrule
\multicolumn{16}{l}{\textbf{Dream Models}} \\
\midrule
Dream-7B-Instruct & 36.8 & 38.0 & 40.2 & 66.9 & 75.2 & 74.4 & 15.0 & 17.5 & 17.5 & 51.2 & 53.6 & 55.4 & 51.8 & 52.2 & 52.8 \\
d1 & \textbf{41.2} & 46.0 & 46.8 & 73.3 & 83.4 & 83.4 & \textbf{20.0} & 15.0 & 15.0 & 54.2 & 53.6 & 54.2 & 53.8 & 54.0 & 53.6 \\
ESPO  & 31.0 & 42.4 & 43.0 & 65.8 & 81.5 & 78.9 & 15.0 & 15.0 & 17.5 & 54.2 & 56.1 & 55.4 & 53.2 & 53.0 & 52.8 \\
TraceRL& 35.2 & 47.8 & 47.6 & 65.9 & 83.6 & 84.6 & 10.0 & 15.0 & 17.5 & 53.6 & 56.7 & 54.8 & 55.2 & 57.0 & 55.4 \\
Coupled-GRPO      & \textbf{41.2} & 45.6 & 45.6 & 69.7 & 82.3 & 82.5 & 10.0 & 20.0 & 25.0 & 54.9 & 54.9 & 57.3 & 54.2 & 54.2 & 54.0 \\
RLDF (ours)          & 39.8 & \textbf{48.6} & \textbf{49.6} & \textbf{74.3} & \textbf{86.5} & \textbf{86.5} & 17.5 & \textbf{30.0} & \textbf{30.0} & \textbf{62.8} & \textbf{61.0} & \textbf{59.2} & \textbf{61.2} & \textbf{60.4} & \textbf{60.6} \\
\rowcolor{gray!15}
$\Delta$ vs. Base & +3.0 & +10.6 & +9.4 & +7.4 & +11.3 & +12.1 & +2.5 & +12.5 & +12.5 & +11.6 & +7.4 & +3.8 & +9.4 & +8.2 & +7.8 \\
\bottomrule
\end{tabular}%
}
\caption{Performance of different RL methods on LLaDA and Dream models across math and code benchmarks under the \textbf{dynamic} unmasking strategy with varying generation steps (128, 256, and 512) with unmasking threshold at 0.9. The best result in each column is highlighted in \textbf{bold}, and $\Delta$ vs. Base denotes the absolute improvement of RLDF (ours) over the corresponding base model.}
\label{tab:main_result}
\end{table*}

\paragraph{Strong Performance Across Tasks and Models.}
Our reinforcement learning from denoising feedback (RLDF) method achieves consistently dominant performance across both LLaDA and Dream models on mathematical reasoning and code generation benchmarks, as reported in Table~\ref{tab:main_result}. Nearly all configurations of our approach rank first or second among all baselines, with first-place results in the majority of settings. Notably, the performance gains are consistent across both model families, in contrast to d1 and ESPO, which demonstrate competitive results only on individual models. Relative to the respective base models, our approach yields substantial improvements of up to 10 accuracy points on Dream for MATH500 and MBPP, and approximately 8 points for LLaDA on the same benchmarks. These empirical results provide compelling evidence for the effectiveness of the proposed method.

\paragraph{Unlocking the Reasoning Potential of the Dream Model.}
An unexpected finding is that the Dream model converges remarkably rapidly during reinforcement learning, requiring only approximately 100 training steps on both math and code tasks. We hypothesize that this unexpectedly fast emergence of reasoning stems from latent capabilities inherited from the underlying Qwen architecture~\citep{yang2025qwen3}, which may have been suppressed during the adaptation from autoregressive to masked diffusion modeling. Our reinforcement learning procedure effectively reactivates these dormant reasoning capacities, recovering and amplifying the model's problem-solving abilities. We further observe that training on Dream exhibits greater instability compared to LLaDA, manifesting as rapidly escalating KL divergence; we address this by applying a higher KL penalty coefficient ($\beta = 0.04$). This instability may reflect a fundamental discontinuity introduced by the autoregressive-to-diffusion adaptation, pointing to a potential limitation of such model conversion approaches.

\paragraph{Cross-dataset generalization.}
Our method generalizes consistently across benchmarks, whereas competing methods often improve on one distribution at the cost of others. Specifically, it yields stable gains on out-of-distribution benchmarks including GSM8K, AMC23, and HumanEval. This is particularly notable for the shift from MBPP to HumanEval: MBPP describes function implementations in free-form natural language, while HumanEval requires completing partially defined function stubs under explicit constraints. The strong cross-dataset performance suggests that our models acquire transferable reasoning skills rather than overfitting to surface patterns of the specific training dataset.

\paragraph{Robustness to unmasking strategy.}
Our main experiments use a dynamic unmasking threshold (set to $0.9$). To further test robustness, we evaluate each model under a static unmasking strategy across varying generation lengths and step counts, constraining the model to unmask one token per step and reporting results at 128, 256, and 512 steps in Table~\ref{tab:static_result}. The static results largely mirror the trends observed under dynamic unmasking, and our method remains dominant particularly on Dream models. These results indicate that the choice of unmasking strategy is a relatively minor factor compared to the number of denoising steps and the generation length.

\begin{table*}[t]
\centering
\renewcommand{\arraystretch}{1.15}
\resizebox{\textwidth}{!}{%
\begin{tabular}{l|ccc|ccc|ccc|ccc|ccc}
\toprule
\multirow{2}{*}{\textbf{Model}} & \multicolumn{3}{c|}{\textbf{MATH500}} & \multicolumn{3}{c|}{\textbf{GSM8K}} & \multicolumn{3}{c|}{\textbf{AMC23}} & \multicolumn{3}{c|}{\textbf{HumanEval}} & \multicolumn{3}{c}{\textbf{MBPP}} \\
\cmidrule(lr){2-4} \cmidrule(lr){5-7} \cmidrule(lr){8-10} \cmidrule(lr){11-13} \cmidrule(lr){14-16}
 & 128 & 256 & 512 & 128 & 256 & 512 & 128 & 256 & 512 & 128 & 256 & 512 & 128 & 256 & 512 \\
\midrule
\multicolumn{16}{l}{\textbf{LLaDA Models}} \\
\midrule
LLaDA-8B-Instruct    & 32.4 & 36.4 & 39.0 & 76.0 & 83.3 & 81.5 & 7.5  & 15.0 & 12.5 & 40.9 & 37.2 & 37.2 & 37.0 & 36.8 & 38.4 \\
LLaDA-1.5            & 31.8 & 40.8 & 39.6 & 76.2 & 82.3 & 83.5 & 15.0 & 12.5 & 10.0 & 43.3 & 36.0 & 39.0 & 36.8 & 38.4 & 38.6 \\
d1                   & 32.2 & 36.8 & 38.8 & 76.0 & 83.0 & 81.7 & 12.5 & \textbf{17.5} & 15.0 & 40.2 & 37.2 & 37.8 & 37.8 & 39.6 & 41.2 \\
ESPO                 & 38.6 & 38.2 & 41.8 & 78.8 & 83.7 & \textbf{86.0} & 10.0 & \textbf{17.5} & 20.0 & 40.9 & 37.8 & 41.5 & \textbf{42.2} & \textbf{45.0} & 42.8 \\
TraceRL              & 32.0 & 33.8 & 33.0 & 77.6 & 80.7 & 82.4 & 7.5  & 5.0  & 20.0 & 40.9 & 34.8 & 37.8 & 36.8 & 39.2 & 40.4 \\
Coupled-GRPO         & 33.0 & 40.6 & 37.0 & 77.9 & 83.5 & 83.6 & 15.0 & 10.0 & \textbf{22.5} & 39.6 & 39.6 & 39.6 & 38.6 & 41.8 & 41.4 \\
RLDF (ours)          & \textbf{40.8} & \textbf{43.6} & \textbf{43.0} & \textbf{79.3} & \textbf{84.6} & 84.5 & \textbf{17.5} & 15.0 & 17.5 & \textbf{43.9} & \textbf{43.3} & \textbf{42.7} & 40.0 & 44.8 & \textbf{45.8} \\
\rowcolor{gray!15}
$\Delta$ vs. Base    & +8.4 & +7.2 & +4.0 & +3.3 & +1.3 & +3.0 & +10.0 & +0.0 & +5.0 & +3.0 & +6.1 & +5.5 & +3.0 & +8.0 & +7.4 \\
\midrule
\multicolumn{16}{l}{\textbf{Dream Models}} \\
\midrule
Dream-7B-Instruct    & 36.2 & 38.4 & 39.2 & 66.2 & 74.9 & 73.9 & 15.0 & 17.5 & 17.5 & 48.2 & 50.0 & 51.2 & 50.8 & 51.0 & 51.8 \\
d1     & \textbf{42.2} & 46.2 & 47.0 & 72.7 & 84.7 & 83.5 & \textbf{20.0} & 15.0 & 15.0 & 53.7 & 51.8 & 51.8 & 52.4 & 53.4 & 53.2 \\
ESPO      & 30.8 & 42.4 & 43.4 & 66.4 & 80.7 & 79.8 & 15.0 & 15.0 & 17.5 & 52.4 & 55.5 & 53.0 & 50.8 & 51.8 & 51.4 \\
TraceRL   & 35.0 & 46.0 & 44.4 & 65.4 & 83.4 & 84.8 & 10.0 & 15.0 & 17.5 & 52.4 & 51.2 & 53.7 & 53.6 & 54.6 & 53.8 \\
Coupled-GRPO         & 40.2 & 43.6 & 45.4 & 70.0 & 82.3 & 82.9 & 12.5 & 17.5 & \textbf{30.0} & 53.0 & 53.0 & 56.1 & 51.8 & 52.2 & 52.2 \\
RLDF (ours)          & 39.4 & \textbf{48.4} & \textbf{49.0} & \textbf{74.0} & \textbf{86.7} & \textbf{86.6} & 17.5 & \textbf{30.0} & \textbf{30.0} & \textbf{58.5} & \textbf{57.9} & \textbf{57.9} & \textbf{58.2} & \textbf{57.2} & \textbf{56.8} \\
\rowcolor{gray!15}
$\Delta$ vs. Base    & +3.2 & +10.0 & +9.8 & +7.8 & +11.8 & +12.7 & +2.5 & +12.5 & +12.5 & +10.4 & +7.9 & +6.7 & +7.4 & +6.2 & +5.0 \\
\bottomrule
\end{tabular}%
}
\caption{Performance of different RL methods on LLaDA and Dream models across math and code benchmarks under the \textbf{static} unmasking strategy with varying generation steps (128, 256, and 512) with one token unmasked each step. The best result in each column is highlighted in \textbf{bold}, and $\Delta$ vs. Base denotes the absolute improvement of RLDF (ours) over the corresponding base model.}
\label{tab:static_result}
\end{table*}

\section{Ablation Studies and Design Choices}

\subsection{Effect of Sampling Numbers and Temperature}

\begin{figure}[!b]
    \centering
    \includegraphics[width=\linewidth]{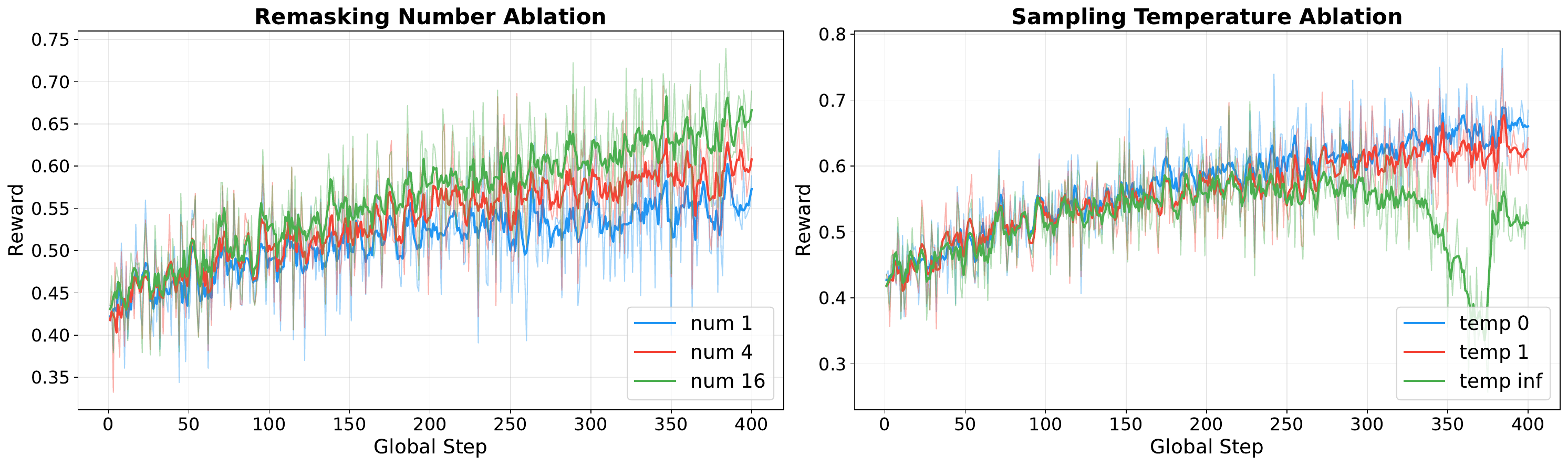}
    \caption{\textbf{Ablation on sequential remasking strategy.} Mean rollout reward under varying remasking counts and timestep sampling temperatures. Larger remasking counts and lower sampling temperatures consistently lead to higher rollout rewards.}
    \label{fig:sampling_ablation}
\end{figure}

We investigate the effect of varying the number of sampling timesteps in Figure~\ref{fig:sampling_ablation}. As expected, increasing the number of sampling timesteps accelerates convergence, analogous to the effect of increasing batch size. Since each additional sampling timestep introduces an extra forward and backward pass, a larger number of timesteps directly increases per-iteration training time. In our implementation, using 16 sampling steps incurs a computation cost comparable to that of the rollout process alone. 

We further observe that the sampling temperature notably affects training stability: lower temperatures yield more stable dynamics. Since lower temperatures bias sampling toward high-entropy timesteps, we attribute this to the loss estimation behavior. High-entropy timesteps are dominated by the few informative tokens that most influence the final generation. In contrast, high temperatures spread sampling toward low-entropy timesteps, whose tokens the model already predicts with high confidence. Repeatedly reinforcing these already-confident predictions drives their probabilities toward saturation and further collapses the entropy, ultimately destabilizing training. This is reflected in Figure~\ref{fig:sampling_ablation}, where the high-temperature run collapses sharply after roughly 300 steps.
% We further observe that the sampling temperature notably affects training stability: lower temperatures yield more stable dynamics. Since lower temperatures bias sampling toward high-entropy timesteps, we attribute this to loss estimation behavior. High-entropy timesteps mainly contain informative tokens, making loss estimation inherently more accurte. Low-entropy timesteps, despite contains more tokens, also bring about less training signals. in contrast, provide far more tokens for stable averaging, though probability-based filtering is needed to isolate informative signal from low-value tokens.

% This reveals a signal-noise trade-off: sufficient signal is required for stable optimization, yet noise from low-entropy timesteps must not dominate the loss estimation. Consequently, choosing an appropriate probability filtering threshold is crucial for training stability.

\subsection{\texorpdfstring{Why $\mathbf{x}_0$ Estimation Outperforms $\mathbf{x}_{t-1}$ Estimation}{Why x_0 Estimation Outperforms x_(t-1) Estimation}}

\begin{figure}[!t]
    \centering
    \includegraphics[width=\linewidth]{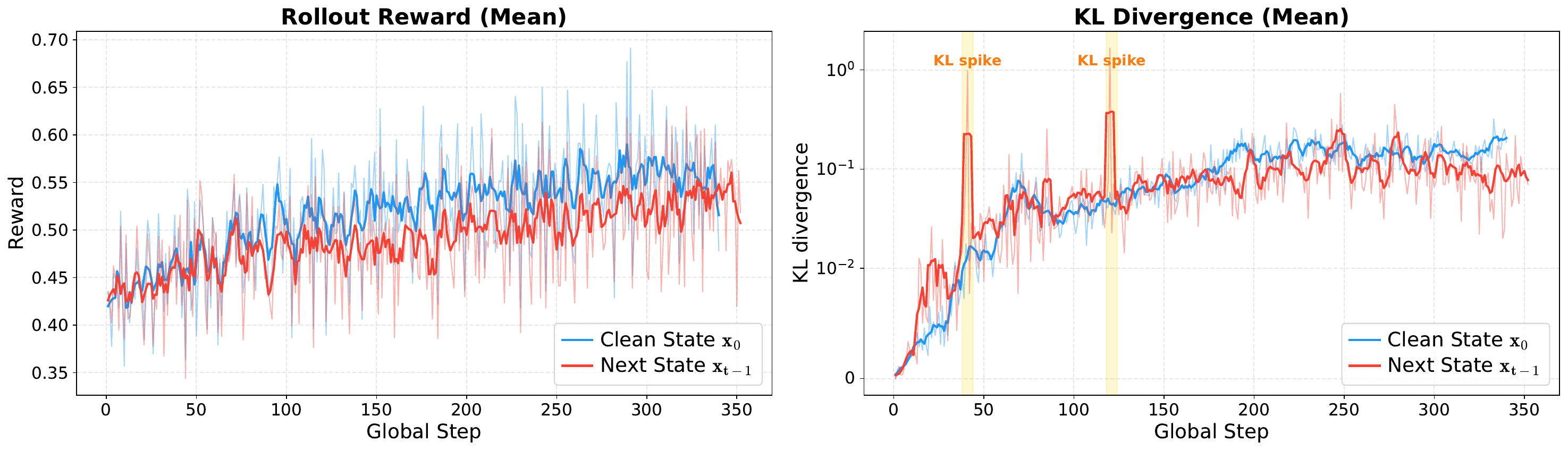}
    \caption{\textbf{Clean state estimation yields more stable training than next state estimation.} Training dynamics on LLaDA code tasks under two target states, measured by rollout reward and KL loss.}
    \label{fig:state_ablation}
\end{figure}

Theoretically, the one-step posterior can be recovered by marginalizing over the clean signal:
\begin{equation}
    p_\theta(x_{t-1} \mid x_t) = \sum_{x_0} p_\theta(x_0 \mid x_t)\, q(x_{t-1} \mid x_t, x_0),
\end{equation}
so any $x_{t-1}$-prediction model implicitly commits to a particular path through $x_0$ space, while $x_0$-prediction retains the full distribution over clean sequences. Predicting $x_0$ thus exposes the model to a global denoising objective at every timestep, yielding a richer signal and a target distribution whose statistics remain stable across noise levels.

This gap is amplified in practice. Under dynamic unmasking, tokens are revealed only when their probability exceeds a threshold (with a single token unmasked otherwise), so the supervisory signal for $x_{t-1}$-estimation is severely limited, especially when high-entropy timesteps are preferentially sampled. As shown empirically in Table~\ref{tab:target_clip}, $x_0$-estimation leverages $\sim$13\% of response tokens per sample, while $x_{t-1}$-estimation uses only $\sim$2\%, discarding most of the available signal.

As shown in Figure~\ref{fig:state_ablation}, $\mathbf{x}_{t-1}$-based estimation also exhibits pronounced KL loss spikes during training, attributable to the instability introduced by computing KL divergence over such a limited token set; the policy loss suffers from the same issue. In contrast, $\mathbf{x}_0$-based estimation yields consistently higher rollout rewards, empirically confirming the superiority of clean state estimation for stable and effective training.

\subsection{Loss Normalization}

Token-level loss normalization has been proposed to mitigate the response-length imbalance commonly observed in RL training of autoregressive models. In this section, we investigate its applicability and effectiveness within our setting. 

Sample-level normalization first normalizes within each sample and then aggregates within the group, ensuring that every sample contributes equally to the batch. In contrast, token-level normalization operates directly over all tokens in the group, giving every token an equal contribution.

\begin{figure}[!t]
    \centering
    \includegraphics[width=\linewidth]{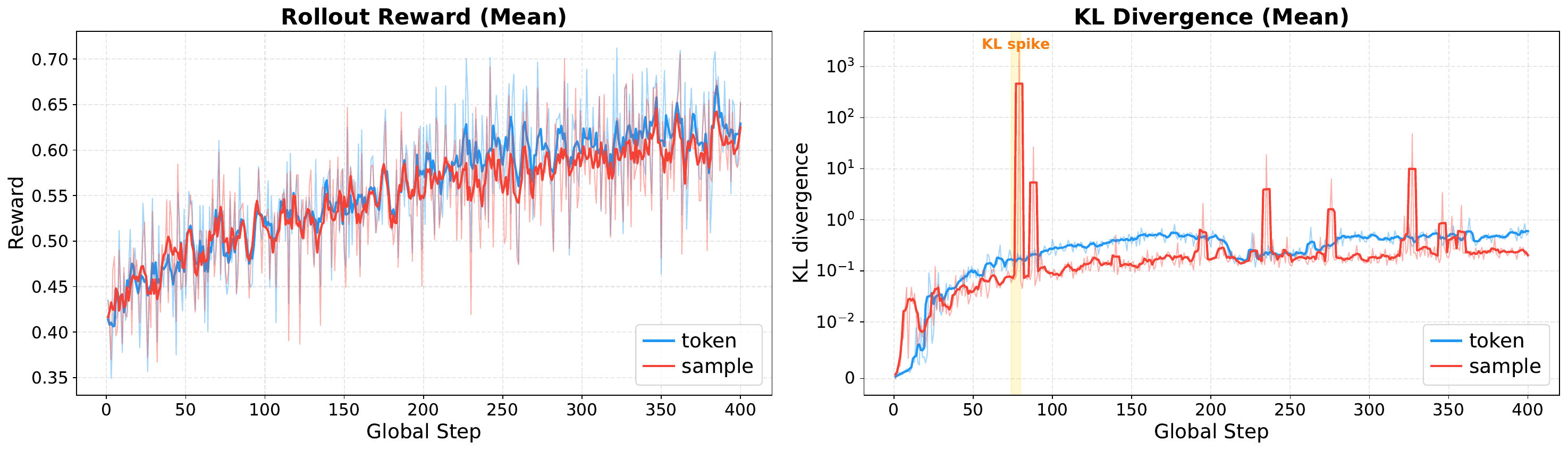}
    \caption{\textbf{Comparison of token-level and sample-level loss normalization (both with gradient clipping).} Token-level normalization leads to a more stable training process, with fewer loss spikes and faster convergence.}
    \label{fig:norm_ablation}
\end{figure}

Formally, in contrast to the sample-level loss defined in Eq.~(\ref{eq:rldf}), the token-level loss is formulated as in Eq.~(\ref{eq:rldf-token}). As shown in Figure~\ref{fig:norm_ablation}, sample-level loss normalization, even when combined with gradient norm clipping, still slightly underperforms its token-level counterpart. Although token-level normalization introduces a deviation from the standard loss objective, it empirically yields a more stable training process by mitigating spikes in KL divergence.
\begin{equation}
\label{eq:rldf-token}
\resizebox{\linewidth}{!}{$
\begin{aligned}
\mathcal{L}_{\text{RLDF}}^{\text{token}}(\theta) = \mathbb{E}_{q\sim\mathcal{D},\,o\sim\pi_{\theta_{\mathrm{old}}}(\cdot\mid q)}\Bigg[
&-\frac{1}{\sum_{b=1}^{G}|\hat{o}^{(b)}_0|}\sum_{b=1}^{G}\sum_{i\in \hat{o}^{(b)}_0}
\mathrm{softmax}_t\bigg(
-\frac{1}{\tau\,|\mathcal{I}_t|}\sum_{j\in \mathcal{I}_t}
\log\phi_{\pi_{\theta_{\mathrm{old}}}}(\hat{o}_{t-1, j}\mid q, o_{t})
\bigg)
\bigg[\min\Big(
\frac{\phi_{\pi_\theta}(\hat{o}_{0,i}^{(b)}\mid q, o_{t}^{(b)})}
     {\phi_{\pi_{\theta_{\mathrm{old}}}}(\hat{o}_{0,i}^{(b)}\mid q, o_{t}^{(b)})}\hat{A}^{(b)},\\[2pt]
& \mathrm{clip}\big(
\frac{\phi_{\pi_\theta}(\hat{o}_{0,i}^{(b)}\mid q, o_{t}^{(b)})}
     {\phi_{\pi_{\theta_{\mathrm{old}}}}(\hat{o}_{0,i}^{(b)}\mid q, o_{t}^{(b)})},\,
1{-}\varepsilon,\,1{+}\varepsilon\big)\hat{A}^{(b)}
\Big)
+\, \beta\, D_{\mathrm{KL}}\Big[\phi_{\pi_\theta}(\cdot\mid q, o_{t}^{(b)})\,\big\|\,\phi_{\pi_{\mathrm{ref}}}(\cdot\mid q, o_{t}^{(b)})\Big]\bigg]
\Bigg]
\end{aligned}
$}
\end{equation}

\subsection{Token Clipping is Necessary}
\setlength{\intextsep}{0pt}
\begin{wraptable}{r}{0.5\columnwidth}
\small
\setlength{\tabcolsep}{4pt}
\renewcommand{\arraystretch}{1.15}
\resizebox{0.5\columnwidth}{!}{%
\begin{tabular}{llccc}
\hline
\multicolumn{2}{c}{\textbf{Setting}} & \multicolumn{3}{c}{\textbf{Training Metrics}} \\
\cmidrule(lr){1-2} \cmidrule(lr){3-5}
\textbf{Target} & \textbf{Clip} & \textbf{Grad Norm} & \textbf{Loss} & \textbf{Token Utility} \\
\hline
$x_0$ & w/   & 8.14       & 0.0727 & 13.39\% \\
$x_0$ & w/o  & 3{,}232.21 & 0.6308 & 62.15\% \\
$x_t$ & N/A    & 8.46       & 0.0591 & 2.07\%  \\
\hline
\end{tabular}}
% \caption{Comparison of training dynamics under different target states and gradient clipping settings. The first two columns specify the experimental \textbf{setting} (target state and whether gradient clipping is applied), while the last three columns report the corresponding \textbf{training metrics} averaged over the first 100 steps. Without token clipping, both gradient norm and loss diverge by several orders of magnitude.}
\caption{Comparison of training dynamics under different target states and gradient clipping settings. The first two columns specify the experimental setting, while the last three columns report the corresponding averaged training metrics. Without token clipping, both gradient norm and loss diverge by several orders of magnitude.}
\label{tab:target_clip}
\end{wraptable}

At each training iteration, the model performs a forward pass conditioned on the partially observed context at timestep $t$. Due to the limited context available, the model struggles to make reliable predictions for positions at timesteps $t' \ll t$, often assigning extremely low probabilities to those tokens. Under the loss formulation in Equation~(\ref{eq:reinforce_loss}), these near-zero probabilities can dominate the loss computation, overshadowing the gradient signal from high-confidence tokens, precisely those from which we seek to learn. This imbalance inflates the noise-to-signal ratio and distorts the training dynamics.

To increase the proportion of informative gradient signal in the policy loss, we apply token-level clipping that discards clean-state tokens whose predicted probability falls below a threshold (0.2 in our implementation). The benefit of this design is validated in Table~\ref{tab:target_clip}: targeting $x_0$ with token clipping achieves a favorable trade-off, yielding a stable gradient norm while substantially improving token utility over the $x_t$ baseline.

\section{Conclusion}

In this work, we address the fundamental challenge of log-likelihood estimation in the policy gradient objective of diffusion language models. We propose \textbf{RLDF} (Reinforcement Learning from Denoising Feedback), a principled method that achieves a favorable trade-off between estimation precision and computational efficiency by approximating the policy loss on clipped final states across a subset of selected timesteps t. Extensive empirical evaluations demonstrate both the effectiveness and the necessity of our approach in enabling stable and performant reinforcement learning for DLMs. Furthermore, we introduce a comprehensive and extensible RL training framework \textbf{Drift} for DLMs, encompassing flexible masking strategies and diverse training methodologies, with native compatibility for the two most prominent DLM architectures, LLaDA and Dream.

\bibliography{colm2026_conference}
\bibliographystyle{colm2026_conference}

\appendix

\section{Classic Variants of Policy Loss}
We introduce three prevalent methods for policy loss estimation, derived from the no masking, random masking, and sequential masking perspectives, respectively.

For brevity, we define the per-token importance ratio as
\begin{equation}
r_{t, i}(\theta) \;=\; \frac{\phi_{\pi_\theta}(o_{0, i} \mid c)}{\phi_{\pi_{\theta_{\text{old}}}}(o_{0, i} \mid c)},
\end{equation}
where the conditioning context $c$ varies across the three variants: 
$c = q$ for no masking, 
$c = (q,\, o^{\mathcal{M}_i})$ for random masking, and 
$c = (q,\, o_{t})$ for sequential masking. 
For all three objectives, $A_i^k$ denotes the token-level advantage, $\varepsilon$ is the PPO clipping threshold, and $\beta$ controls the strength of the KL regularization toward the reference policy $\pi_{\text{ref}}$. By convention, quantities without an explicit subscript $t$ correspond to the final step $t = T$ (e.g., $r_i \triangleq r_{0, i}$, $o_i \triangleq o_{0, i}$).

% Diffu-GRPO
\paragraph{All Masking.} 
Diffu-GRPO~\cite{zhao2025d1} computes the policy-gradient loss over \emph{all} tokens of every sampled response. For each query $q$, a randomly masked variant $q'$ is constructed and used as the conditioning context, providing a one-step likelihood estimation at every position $i \in \{1,\dots,|o^{(b)}|\}$:
\begin{equation}
\resizebox{\linewidth}{!}{$
\displaystyle
\mathcal{L}_{\text{diffu}}(\theta) = \mathbb{E}\Bigg[ \frac{1}{G}\sum_{b=1}^{G}\frac{1}{|o^{(b)}|}\sum_{i=1}^{|o^{(b)}|}\min\Big( r_i^{(b)} A^{(b)},\ \text{clip}(r_i^{(b)}, 1-\varepsilon, 1+\varepsilon)\, A^{(b)} \Big) - \beta\, \mathbb{D}_{\text{KL}}\big[\phi_{\pi_\theta}\,\|\,\phi_{\pi_{\text{ref}}}\big]\Bigg]
$}
\end{equation}
The expectation is taken over $q \sim \mathcal{D}$, and $\{o^{(b)}\}_{b=1}^{G} \sim \pi_{\theta_{\text{old}}}(\cdot \mid q)$. Because the loss is averaged uniformly over $|o^{(b)}|$ tokens, every output position contributes equally regardless of whether it was actually denoised at that step.

% ELBO-GRPO
\paragraph{Random Masking.} 
Instead of scoring all tokens, random masking restricts the loss to a randomly sampled mask set $\mathcal{M}_b \subseteq \{1,\dots,|o^{(b)}|\}$ for each response, while the unmasked tokens $o^{{\overline{\mathcal{M}_b}}}$ and prompt $q$ serve as the context. This corresponds to a Monte-Carlo estimator of the masked-diffusion ELBO at the token level:
\begin{equation}
\resizebox{\linewidth}{!}{$
\displaystyle
\mathcal{L}_{\text{ELBO}}(\theta) = \mathbb{E}\Bigg[ \frac{1}{G}\sum_{b=1}^{G}\frac{1}{|\mathcal{M}_b|}\sum_{i \in \mathcal{M}_b}\min\Big( r_i^{(b)} A^{(b)},\ \text{clip}(r_i^{(b)}, 1-\varepsilon, 1+\varepsilon)\, A^{(b)} \Big) - \beta\, \mathbb{D}_{\text{KL}}\big[\phi_{\pi_\theta}\,\|\,\phi_{\pi_{\text{ref}}}\big]\Bigg]
$}
\end{equation}
The expectation is taken over $q \sim \mathcal{D}$, $\{o^{(b)}\}_{b=1}^{G} \sim \pi_{\theta_{\text{old}}}(\cdot \mid q)$, and $\mathcal{M}_b \sim \text{masking}(o^{(b)})$. By conditioning on the visible tokens and supervising only the masked positions, ELBO yields a lower-variance, theoretically grounded surrogate that is consistent with the training objective of masked diffusion language models. ESPO~\cite{ou2025principled} is modified from the ELBO estimation.

% Trajectory-GRPO
\paragraph{Sequential Masking.} 
The sequential masking method further aligns the RL signal with the actual generation process by following the diffusion denoising trajectory $\tau = (o_T, o_{T-1}, \dots, o_0)$, where $o_T$ denotes the fully masked initial state and $o_0 = o$ is the final clean response. At each denoising timestep $t$, the policy unmasks a subset of positions $\mathcal{I}_t$ conditioned on the partially denoised state $o_{t}$, producing the next state $o_{t-1}$. The loss is computed at exactly those positions:
\begin{equation}
\resizebox{\linewidth}{!}{$
\displaystyle
\mathcal{L}_{\tau}(\theta) = \mathbb{E}\Bigg[ \frac{1}{G}\sum_{b=1}^{G}\frac{1}{T}\sum_{t=1}^{T}\frac{1}{|\mathcal{I}^{(b)}_t|}\sum_{i \in \mathcal{I}^{(b)}_t}\min\Big( r_{t, i}^{(b)} A^{(b)},\ \text{clip}(r_{t, i}^{(b)}, 1-\varepsilon, 1+\varepsilon)\, A^{(b)} \Big) - \beta\, \mathbb{D}_{\text{KL}}\big[\phi_{\pi_\theta}\,\|\,\phi_{\pi_{\text{ref}}}\big]\Bigg]
$}
\end{equation}
The expectation is taken over $q \sim \mathcal{D}$ and $o_{t-1} \sim \pi_{\theta_{\text{old}}}(\cdot \mid q,\, o_{t})$ for $t = T, T-1, \dots, 1$ and $b = 1, \dots, G$. Unlike the previous two variants, every token is supervised under the exact context in which it was generated, so the optimization landscape matches the inference-time denoising procedure step by step. This faithfulness comes at the cost of $T$-fold more forward passes per response. \textbf{Our method adapts the sequential masking variant by predicting the clean state $\mathbf{x}_0$ instead of $\mathbf{x}_{t-1}$, retaining its trajectory-level precision at a fraction of the computational cost, thus striking a favorable balance between fidelity and efficiency.}

\section{Reward Function}
\label{app:reward}

For \textbf{math tasks}, we extract the content of the last $\backslash$\texttt{boxed\{...\}} from the model output and compare it against the ground truth via a math-equivalence check (combining string normalization, \texttt{sympy}-based symbolic comparison, and the \texttt{math\_verify} library). The reward is binary:
\begin{equation}
r_{\mathrm{math}} = 
\begin{cases}
1, & \text{if } \hat{a} \equiv a^* \\
0, & \text{otherwise}
\end{cases}\nonumber
\end{equation}
where $\hat{a}$ is the extracted answer and $a^*$ the ground truth.

For \textbf{code tasks}, we extract the Python code block enclosed by \verb|```python ```| and execute it against the provided unit tests in a sandboxed runner. The reward is the test-case pass rate:
\begin{equation}
r_{\mathrm{code}} = \frac{\#\,\text{passed test cases}}{\#\,\text{total test cases}} \in [0, 1],\nonumber
\end{equation}
yielding a dense signal that rewards partial correctness and provides smoother gradients than the binary math reward.

\section{Implementation Details}
\label{app:impl}
\subsection{Data Processing}
Out-of-memory (OOM) errors frequently occur during both rollout and training. To mitigate this, we restrict training data to subsets of the MATH and MBPP training splits with a maximum sequence length of 320 tokens for LLaDA and 256 tokens for Dream. Additionally, since binary rewards on MATH yield limited advantage diversity within each group, we retain only samples for which LLaDA produces at least one correct response across 16 generations, ensuring sufficient richness in the reward signal.
\subsection{Remasking Strategy}
We adopt a sequential remasking strategy with weighted timestep sampling. The sampling temperature defined in Equation~\ref{eq:step_softmax} is set at one. To balance training and rollout efficiency, we set k=16, meaning 16 timesteps are sampled per training example.
\subsection{Training}
We use separate batch sizes for rollout and training. Prior to each training update, rollout responses are filtered to retain only samples exhibiting non-trivial policy gradients, i.e., those with varying reward values within a group. We apply a REINFORCE loss with a single gradient update per sample, rather than PPO. The KL penalty coefficient $\beta$ is set to 0.01 for LLaDA and 0.04 for Dream. The learning rate is fixed at $1e^{-6}$ throughout training.
\subsection{Rollout}
A dynamic unmasking strategy is used for both rollout and evaluation. Tokens are unmasked greedily by highest probability when no token exceeds the confidence threshold of 0.9. To accelerate generation, we employ block-wise denoising with a block size of 32. LLaDA samples via the Gumbel-Max trick, adding Gumbel noise to logits and taking the argmax to implicitly approximate softmax sampling, with temperature controlling noise magnitude. Dream instead uses explicit categorical sampling, applying temperature scaling with optional top-p/top-k truncation before sampling from the softmax distribution. The rollout batch size is set to 8, with 4 generations per sample.

\section{Planning}
To broaden our evaluation, we apply each method to planning tasks, adopting the data-processing pipeline of \citet{zhao2025d1}. Table~\ref{tab:planning_result} reports results on the Countdown and Sudoku planning tasks across two diffusion-LM backbones (LLaDA and Dream) and both unmasking schedules. RLDF improves over the corresponding base model in all settings, with gains ranging from 34 to 71 points. These improvements are consistent across both backbones, whereas prior RL methods are markedly backbone-dependent. ESPO and Coupled-GRPO are competitive on LLaDA but degrade substantially on Dream, while d1 exhibits the opposite pattern. RLDF is the only method that remains strong on both backbones. We note that Countdown and Sudoku are highly structured and closely tied to their training distributions, which leaves room for solutions that exploit task-specific regularities rather than reflecting general reasoning ability. We therefore analyze mathematics and code in detail in our main experiments (Section~\ref{sec:main_results}).

\begin{table*}[!htbp]
\centering
\renewcommand{\arraystretch}{1.15}
\resizebox{\textwidth}{!}{%
\begin{tabular}{l|cccccccc|cccccccc}
\toprule
\multirow{4}{*}{\textbf{Method}} & \multicolumn{8}{c|}{\textbf{LLaDA}} & \multicolumn{8}{c}{\textbf{Dream}} \\
\cmidrule(lr){2-9} \cmidrule(lr){10-17}
 & \multicolumn{4}{c|}{\textbf{Dynamic}} & \multicolumn{4}{c|}{\textbf{Static}}
 & \multicolumn{4}{c|}{\textbf{Dynamic}} & \multicolumn{4}{c}{\textbf{Static}} \\
\cmidrule(lr){2-5}\cmidrule(lr){6-9}\cmidrule(lr){10-13}\cmidrule(lr){14-17}
 & \multicolumn{2}{c}{\textbf{Countdown}} & \multicolumn{2}{c|}{\textbf{Sudoku}}
 & \multicolumn{2}{c}{\textbf{Countdown}} & \multicolumn{2}{c|}{\textbf{Sudoku}}
 & \multicolumn{2}{c}{\textbf{Countdown}} & \multicolumn{2}{c|}{\textbf{Sudoku}}
 & \multicolumn{2}{c}{\textbf{Countdown}} & \multicolumn{2}{c}{\textbf{Sudoku}} \\
\cmidrule(lr){2-3}\cmidrule(lr){4-5}\cmidrule(lr){6-7}\cmidrule(lr){8-9}\cmidrule(lr){10-11}\cmidrule(lr){12-13}\cmidrule(lr){14-15}\cmidrule(lr){16-17}
 & 256 & 512 & 256 & 512 & 256 & 512 & 256 & 512 & 256 & 512 & 256 & 512 & 256 & 512 & 256 & 512 \\
\midrule
Base (Instruct) & 26.2 & 21.5 & 9.2 & 2.2 & 26.6 & 22.7 & 8.6 & 2.2 & 27.7 & 25.8 & 41.8 & 31.4 & 30.1 & 27.0 & 26.6 & 22.0 \\
d1            & 38.3 & 30.1 & 8.2 & 2.2 & 37.1 & 30.5 & 8.4 & 2.2 & 68.8 & 58.6 & 63.4 & 48.2 & \textbf{69.1} & 60.9 & 51.4 & 58.6 \\
ESPO         & 70.3 & 64.1 & 81.8 & 55.0 & 71.1 & 64.1 & 83.0 & 61.4 & 29.7 & 19.9 & 59.2 & 65.8 & 26.6 & 19.1 & 55.2 & 61.8 \\
Coupled-GRPO & 58.6 & 52.3 & \textbf{86.6} & \textbf{72.0} & 57.0 & 51.2 & \textbf{86.6} & \textbf{71.6} & 32.8 & 29.7 & 35.6 & 41.6 & 33.6 & 27.7 & 42.4 & 37.0 \\
RLDF (ours)  & \textbf{75.0} & \textbf{74.6} & 80.4 & \textbf{72.0} & \textbf{75.0} & \textbf{75.0} & 80.2 & 71.2 & \textbf{69.5} & \textbf{67.2} & \textbf{81.2} & \textbf{82.6} & 64.8 & \textbf{64.1} & \textbf{78.0} & \textbf{82.6} \\
\rowcolor{gray!15}
$\Delta$ vs. Base & +48.8 & +53.1 & +71.2 & +69.8 & +48.4 & +52.3 & +71.6 & +69.0 & +41.8 & +41.4 & +39.4 & +51.2 & +34.7 & +37.1 & +51.4 & +60.6 \\
\bottomrule
\end{tabular}%
}
\caption{Performance of RL methods on \textbf{LLaDA} and \textbf{Dream} under dynamic and static unmasking, on Countdown (Countdown) and Sudoku with 256/512 generation steps. ``Base (Instruct)'' is LLaDA-8B-Instruct / Dream-7B-Instruct respectively. Best per column in \textbf{bold}; $\Delta$ vs. Base is RLDF over the corresponding base.}
\label{tab:planning_result}
\end{table*}

\section{Prompt}
\paragraph{Math Prompt\\}
\begin{tcolorbox}[
    colback=gray!5,
    colframe=gray!50,
    boxrule=0.5pt,
    arc=2pt,
    left=8pt, right=8pt, top=6pt, bottom=6pt
]
\ttfamily
You need to put your final answer in \textbackslash boxed\{\}. This is the problem:\\
\{\{problem\}\}
\end{tcolorbox}

\paragraph{Code Prompt\\}
\begin{tcolorbox}[
    colback=gray!5,
    colframe=gray!50,
    boxrule=0.5pt,
    arc=2pt,
    left=8pt, right=8pt, top=6pt, bottom=6pt
]
\ttfamily
\{\{problem\}\}\\
Place your code within a single Python code block \verb|```python ```|.\\
Do not include more than one code block.
\end{tcolorbox}

\paragraph{Countdown Prompt\\}
\begin{tcolorbox}[
    colback=gray!5,
    colframe=gray!50,
    boxrule=0.5pt,
    arc=2pt,
    left=8pt, right=8pt, top=6pt, bottom=6pt
]
\ttfamily
Please solve the following Countdown puzzle.\\
\\
You are given 3 numbers: \{\{numbers\}\}. Your goal is to use these numbers with basic arithmetic operations (+, -, *, /) to reach the target number \{\{target\}\}.\\
\\
Rules:\\
- You must use each number at most once.\\
- You can use +, -, *, / and parentheses.\\
- Intermediate results must be positive integers (no fractions, no negative numbers).\\
- You do not need to use all the numbers.\\
\\
Show your step-by-step reasoning, then output the full expression that equals the target number inside \textbackslash boxed\{\}.\\
\\
For example, if the numbers are 5, 3, 2 and the target is 11, your answer would be \textbackslash boxed\{5 + 3 * 2\}.\\
\\
Now solve this puzzle:\\
Numbers: \{\{numbers\}\}\\
Target: \{\{target\}\}
\end{tcolorbox}

\paragraph{Sudoku Prompt\\}
\begin{tcolorbox}[
    colback=gray!5,
    colframe=gray!50,
    boxrule=0.5pt,
    arc=2pt,
    left=8pt, right=8pt, top=6pt, bottom=6pt
]
\ttfamily
Please solve the following $4\times4$ Sudoku puzzle. The puzzle is provided as a $4\times4$ grid where `0' represents empty cells.\\
\\
\textbf{Rules:}\\
- Fill empty cells with digits 1--4.\\
- Each row must contain digits 1--4 exactly once.\\
- Each column must contain digits 1--4 exactly once.\\
- Each $2\times2$ box must contain digits 1--4 exactly once.\\
\\
\textbf{Example:}\\
Puzzle:\\
0 4 0 1\\
0 0 2 0\\
1 0 0 3\\
0 3 1 0\\
\\
Solution:\\
2 4 3 1\\
3 1 2 4\\
1 2 4 3\\
4 3 1 2\\
\\
\textbf{Important:} Your solution must be a complete $4\times4$ grid using only digits 1--4, with each row on a new line and digits separated by single spaces.\\
\\
Respond in this exact format:\\
\textless reasoning\textgreater\\
Your step-by-step solving process\\
\textless /reasoning\textgreater\\
\textless answer\textgreater\\
row1\\
row2\\
row3\\
row4\\
\textless /answer\textgreater\\
\\
Now solve this puzzle:\\
Puzzle:\\
\{\{puzzle\}\}
\end{tcolorbox}

\section{Algorithm}
We present the full procedure of RLDF in the algorithms below, organized as a main loop (Algorithm~\ref{alg:rldf}) together with two subroutines: a weighted timestep sampling function (Algorithm~\ref{alg:sample-steps}) and a per-timestep loss estimation (Algorithm~\ref{alg:step-loss}). The main loop follows a GRPO-style group-relative scheme: for each sample, it draws a group of
$G$ responses from the old policy, normalizes their rewards into advantages, and updates the policy over
$N$ inner iterations. Since a full denoising trajectory contains far more timesteps than is practical to optimize over, the main loop calls \textsc{SampleSteps} to select an informative subset of $k$ timesteps for each response, and \textsc{TimestepLoss} to compute the objective at each sampled timestep. We describe each subroutine in turn below.
\begin{algorithm*}[!htbp]
\caption{Reinforcement Learning from Denoising Feedback}
\begin{algorithmic}[1]
\Require Policy $\pi_\theta$, reference $\pi_{\text{ref}}$, dataset $\mathcal{D}$;
group size $G$, inner iterations $N$, sampled steps $k$;
clip range $(\epsilon_{\text{lo}}, \epsilon_{\text{hi}})$, KL weight $\beta$,
learning rate $\eta$, std floor $\delta$, sampling temperature $\tau$
\For{each sample $(q, y) \sim \mathcal{D}$}
  \State $\theta_{\text{old}} \gets \theta$
  \For{$b = 1, \ldots, G$}
    \State Sample response $o^{(b)} \sim \pi_{\theta_{\text{old}}}(\cdot \mid q)$;
           record step probs $u$ and denoising trajectories $\mathcal{T}$
  \EndFor
  \State $r^{(b)} \gets R(o^{(b)}, y)$;\quad
         $\mu \gets \tfrac{1}{G}\sum_b r^{(b)}$;\quad
         $\sigma \gets \sqrt{\tfrac{1}{G}\sum_b (r^{(b)} - \mu)^2}$
  \State $A^{(b)} \gets \big[(r^{(b)} - \mu) / \max(\sigma, \delta)\big]$
  \For{$b = 1, \ldots, G$}
    \State $\mathcal{S}^{(b)} \gets \textsc{SampleSteps}(u, \tau)$
      \Comment{See Algorithm~\ref{alg:sample-steps}}
  \EndFor
    \For{$n = 1, \ldots, N$}
      \State $\nabla_\theta \mathcal{L} \gets 0$
      \For{$b = 1, \ldots, G$}
        \For{$t \in \mathcal{S}^{(b)}$}
          \State Apply masking to $o^{(b)}$ at step $t$ to obtain denoising state $o_t^{(b)}$
          \State $\nabla_\theta \mathcal{L} \mathrel{+}= \tfrac{1}{Gk}\,
                  \nabla_\theta\, \textsc{TimestepLoss}(o_0^{(b)}, o_t^{(b)})$
            \Comment{See Algorithm~\ref{alg:step-loss}}
        \EndFor
      \EndFor
      \State $\theta \gets \theta - \eta \cdot \nabla_\theta \mathcal{L}$
    \EndFor
  \EndFor
\end{algorithmic}
\label{alg:rldf}
\end{algorithm*}

\begin{algorithm*}[!htbp]
\caption{Weighted Timestep Sampling}
\begin{algorithmic}[1]
\Require Recorded step probabilities $u$
         from rollout; sampling temperature $\tau$;
         number of steps to sample $k$;
         trajectories $\mathcal{T}$
\Function{SampleSteps}{$u, \tau, k$}
  \State $w_t \leftarrow \dfrac{\exp(u_t/\tau)}{\sum_{t' \in \mathcal{T}} \exp(u_{t'}/\tau)}$ \quad for each timestep $t \in \mathcal{T}$
  \State Sample $k$ steps $\mathcal{S} \subset \mathcal{T}$ without replacement from $\{w_t\}$
  \State \Return sampled step set $\mathcal{S}$
\EndFunction
\end{algorithmic}
\label{alg:sample-steps}
\end{algorithm*}

\begin{algorithm*}[!htbp]
\caption{Loss Estimation of Each Timestep}
\begin{algorithmic}[1]
\Require Query $q$;
         original response $o_0$ and intermediate state $o_t$;
         current policy $\pi_\theta$, old policy $\pi_{\theta_{\mathrm{old}}}$,
         reference policy $\pi_{\mathrm{ref}}$;
         KL weight $\beta$;
         inner iterations $N$
\Function{TimestepLoss}{$o_0, o_t$}
    \State Clip $o_0$ to obtain $\hat{o}_0$ by thresholding
           on $\pi_{\theta}(o_0 \mid q, o_t)$
  \If{$N = 1$} \Comment{REINFORCE}
    \State $\ell_{\text{policy}} = -\dfrac{1}{|\hat{o}_0|}\,
            \log \pi_{\theta}(\hat{o}_{0} \mid q,\, o_{t})\,\hat{A}$
  \Else \Comment{PPO}
    \State $r_{t}(\theta) = \phi_{\pi_\theta}(\hat{o}_{0}\mid q, o_{t})
            \,/\, \phi_{\pi_{\theta_{\mathrm{old}}}}(\hat{o}_{0}\mid q, o_{t})$
    \State $\ell_{\text{policy}} = \min\!\Big(
            r_{t}\hat{A},\;
            \mathrm{clip}(r_{t},\,1{-}\varepsilon,\,1{+}\varepsilon)\,\hat{A}
            \Big)$
  \EndIf
  \State $\ell_{\text{KL}} = D_{\mathrm{KL}}\!\left[\phi_{\pi_\theta}(\cdot\mid q, o_{t})
            \,\big\|\,\phi_{\pi_{\mathrm{ref}}}(\cdot\mid q, o_{t})\right]$
  \State $\ell(t) = \ell_{\text{policy}} + \beta\, \ell_{\text{KL}}$
  \State \Return $\ell(t)$
\EndFunction
\end{algorithmic}
\label{alg:step-loss}
\end{algorithm*}

\end{document}